\documentclass[conference]{IEEEtran}
\usepackage{amsmath, amssymb, graphicx}
\usepackage{cite}

\usepackage{float}
\usepackage{caption}
\usepackage{subcaption}
\usepackage{xcolor}
\usepackage{hyperref}
\usepackage{enumitem}
\usepackage{mathtools}

\usepackage{siunitx}

\usepackage[normalem]{ulem}

\def\x{{\mathbf x}}

\DeclareMathOperator{\E}{\mathbb{E}}

\usepackage[colorinlistoftodos]{todonotes}

\colorlet{LightTeal}{white!70!teal}
\colorlet{LightOrange}{white!70!orange}

\definecolor{DarkGreen}{RGB}{1,100,32}

\input{mysymbol.sty}

\def\ninept{\def\baselinestretch{1.0}\let\normalsize\small\normalsize}

\title{Generative Diffusion Models for Resource Allocation in Wireless Networks}
\author{
    \IEEEauthorblockN{Yi\u{g}it~Berkay~Uslu,
    Samar~Hadou,
    Shirin~Saeedi~Bidokhti,
    Alejandro~Ribeiro
    }
    \IEEEauthorblockA{
    University of Pennsylvania
    \\\{ybuslu, selaraby, saeedi, aribeiro\}@seas.upenn.edu}
}

\DeclareMathOperator*{\maximum}{\text{maximum}}

\renewcommand {\st}{\text{\,subject to}~}

\newtheorem{problem}{\hspace{0pt}\bf Problem}

\begin{document}
\ninept
\maketitle

\begin{abstract}
    This paper proposes a supervised training algorithm for learning stochastic resource allocation policies with generative diffusion models (GDMs). We formulate the allocation problem as the maximization of an ergodic utility function subject to ergodic Quality of Service (QoS) constraints.
    Given samples from a stochastic expert policy that yields a near-optimal solution to the constrained optimization problem, we train a GDM policy to imitate the expert and generate new samples from the optimal distribution. We achieve near-optimal performance through the sequential execution of the generated samples. To enable generalization to a family of network configurations, we parameterize the backward diffusion process with a graph neural network (GNN) architecture. We present numerical results in a case study of power control.
\end{abstract}


\vspace{1em}
\section{Introduction}
\label{sec:intro}

Most existing formulations and methods for optimal wireless resource allocation, whether classical or learning-based, seek deterministic solutions. In contrast, \emph{optimal solutions} of many non-convex optimization problems (e.g., power control, scheduling) are \emph{inherently probabilistic}, as the optimal solution may lie in the convex hull of multiple deterministic policies. By randomizing between multiple deterministic strategies, stochastic policies can achieve better performance by effectively convexifying the problem~\cite{neely2010stochastic}. This phenomenon is also fundamental in multi-user information theory, where time sharing plays a critical role in achieving optimal performance across various communication channels \cite{gallager1985multiaccess, sason2004achievablerate,He2020}. In this work, we leverage diffusion models to learn generative solutions to stochastic network resource allocation problems.

Generative models (GMs) have shown significant success in generating samples from complex, multi-modal data distributions. Among the wide class of generative models
including variational autoencoders (VAEs) and generative adversarial networks (GANs), \emph{generative diffusion models} (GDMs) stand out for their capability of generating high-quality and diverse samples with stable training \cite{ho2020denoising, rombach2020highresolution}. GDMs convert target data samples (e.g., images) to samples from an easy-to-sample prior (e.g., isotropic Gaussian noise) by a forward (noising) process, and then learn a backward (denoising) process to transform the prior distribution back to the target data distribution.   

A substantial body of the existing literature utilizes GDMs, and GMs in general, for generating domain-specific synthetic data and for data augmentation to enhance the machine-learning models in supervised and reinforcement learning tasks
\cite{kasgari2020experienced, njima2022dnn}.
Yet, research on the use of GMs for wireless network optimization, and GDMs in particular, is scant \cite{Diallo2024GenerativeNetworks, hua2019gan, Du2024EnhancingOptimization, nouri2025diffusion, you2025dress}. 
Concurrent works~\cite{diffsg2024liang, Liang2024DiffusionModelsNetworkOptimizers,  darabi2024diffusion, Lu25Energy, xue2025joint, wang2025graph} propose generative model solvers for network optimization as a framework to learn solution distributions that concentrate the probability mass around optimal deterministic solutions. The generative process then converts random noise to high-quality solutions by eliminating the noise introduced in the forward process.
However, the problem formulation in the aforementioned studies is deterministic and ignores the probabilistic nature of the optimal solution.

Our work is one of the first to imitate \emph{stochastic} expert policies using GDMs. We emphasize the stochastic nature of certain network optimization problems where random solutions are 
not only essential for optimality but also are realized by leveraging iterative dual domain algorithms. In our approach, Quality of Service (QoS) near-optimality emerges through the sequential execution of solutions sampled from the optimal generated distribution.  Moreover, we use a graph neural network (GNN) architecture as the backbone for the reverse diffusion process to enable learning families of solutions across network topologies. 
GNNs not only excel in learning policies from graph-structured data \cite{shen2020graph,wang2022decentralized, StateAugmented_RRM_GNN_naderializadeh_TSP2022, uslu2025faststateaugmentedlearningwireless} but also exhibit desirable properties such as stability, transferability and permutation-equivariance \cite{ruiz2021gnnstability, testa2024stability}.

This paper tackles imitation learning of stochastic wireless resource allocation policies. A GDM policy is trained to match an optimal solution distribution to a constrained optimization problem from which an expert policy can sample (Section~\ref{sec:problem_formulation} and Section~\ref{sec:proposed-gdm-policies}). We utilize a GNN-parametrization to condition the generative diffusion process directly on the network graphs (Section~\ref{sec:gnn-architectures}). We evaluate the proposed GDM policy in a power control setup and demonstrate that the trained GDM policy closely matches the expert policy over a family of wireless networks (Section~\ref{sec:experiments}).


\section{Optimal Wireless Resource Allocation} \label{sec:problem_formulation}

We represent the channel state of a wireless (network) system with a matrix $\h \in \ccalH \subseteq \reals^{N \times N}$ and the allocation of corresponding resources with a vector $\x \in \ccalX \subseteq \reals^N$. Given $\h$, the choice of resource allocation $\x$ determines several QoS metrics that we represent with an objective utility $f_0: \ccalX \times \ccalH \mapsto \reals$ and a constraint utility $\bbf: \ccalX \times \ccalH \mapsto \reals^c$. We define an optimal resource allocation $\bbx^*(\h)$ as the argument that solves the constrained optimization problem,
\begin{alignat}{3} \label{problem:deterministic-functional}
    \tilde{\text{P}}(\h) 
        = f_0\big(\bbx^*(\h), \h \big) 
        = &\maximum_{\x \in \ccalX} ~
              && f_0 \big( \x(\h), \h \big), \nonumber \\
          &\st ~
              && \bbf \big( \x(\h), \h \big) \geq \bb0.
\end{alignat}
In \eqref{problem:deterministic-functional}, we seek a resource allocation $\bbx^*(\h)$ with the largest $f_0$ utility among those in which the components of the utility $\bbf$ are nonnegative. This abstract formulation encompasses channel and power allocation \cite{liang2019towards} in wireless networks (Section \ref{sec:experiments}) as well as analogous problems, in, e.g., point-to-point \cite{yu2006sumcapacity}, MIMO \cite{lin2015massivemimo}, broadcast \cite{tse1997optimal} and interference channels \cite{Chaitanya13interference}.

In most cases of interest, the utilities $f_0$ and $\bbf$ in \eqref{problem:deterministic-functional} are not convex. For this reason, we introduce the convex relaxation in which optimization is over probability {distributions} of resource allocation variables and QoS is measured in {expectation}, 
\begin{alignat}{3} \label{problem:stochastic-functional}
    {\text{P}}(\h) 
        ~=~ &\maximum_{\Dx} ~
              && \E_{\Dx} \Big[\,f_0 \big( \x(\h), \h \big) \, \Big], \nonumber \\
          &\st ~
              && \E_{\Dx} \Big[\, \bbf \big( \x(\h), \h \big) \,\Big]\geq \bb0.
\end{alignat}
In \eqref{problem:stochastic-functional}, we search over stochastic policies $\Dx$ that maximize the \emph{expected} utility $\E_{\Dx} [\,f_0 ( \x(\h), \h )]$ while satisfying the \emph{expected} constraint $\E_{\Dx} [\bbf \big( \x(\h), \h)]\geq \bb0$ when the resource allocation $\x(\h)$ is drawn from the distribution $\Dx$. For future reference, we introduce $\Dx^*(\h) = \Dx^*(\bbx\given\h)$ to denote a distribution that solves \eqref{problem:stochastic-functional}. In $\Dx^\star(\h)$, the channel state $\h$ is given and allocations $\x$ are sampled. 

The important point here is that the performance of stochastic policies is realizable through time sharing if we allocate resources in a faster time scale than QoS perception. Indeed, if we consider independent resource allocation policies  $\x_{\tau}(\h) \sim \Dx$ we have that for sufficiently large $T$,
\begin{alignat}{3} \label{eqn_time_sharing}
    \frac{1}{T} \sum_{\tau=1}^T f_0 \big( \x_{\tau}(\h), \h \big)
        \approx
            \E_{\Dx} \Big[\,f_0 \big( \x(\h), \h \big) \, \Big],
\end{alignat}
with an analogous statement holding for the constraint utility $\bbf$. Since deterministic policies are particular cases of stochastic policies, we know that ${\text{P}}(\h) \geq \tilde{\text{P}}(\h)$. In practice, it is often the case that ${\text{P}}(\h) \gg \tilde{\text{P}}(\h)$ and for this reason, the stochastic formulation in \eqref{problem:stochastic-functional} is most often preferred over the deterministic formulation in \eqref{problem:deterministic-functional}, \cite{neely2010stochastic, gallager1985multiaccess, sason2004achievablerate}.


\subsection{Imitation Learning of Stochastic Policies}

In this paper, we want to learn to imitate the stochastic policies that solve \eqref{problem:stochastic-functional}. Consider a distribution $\Dh$ of channel states $\h$. For each realization $\h$, recall that the solution of \eqref{problem:stochastic-functional} is the probability distribution $\Dx^*(\h)=\Dx^*(\bbx\given\h)$. Separate from these optimal distributions, we consider a parametric family of conditional distributions $\Dx(\h; \bbtheta) = \Dx (\x \given \h; \bbtheta)$ in which the channel state $\h$ is given, and resource allocation variables are drawn. Our goal is to find the conditional distribution $\Dx^*(\h; \bbtheta)$ that minimizes the expectation of the conditional KL-divergences $\kldiv[\big]{\Dx^*(\h)}{\Dx(\h; \bbtheta)}$,
\begin{equation}\label{problem:supervised-diffusion-learning}
  \Dx^*(\h; \bbtheta)
     \,=\, \argmin_{\Dx(\h; \bbtheta)} ~
           \E_{\Dh} 
               \Big[\, \kldiv[\big]{\Dx^*(\h)}{\Dx(\h; \bbtheta)} \,\Big],
\end{equation}
In \eqref{problem:supervised-diffusion-learning}, the distributions $\Dx^*(\h)$ are given for all $\h$. The conditional distribution $\Dx(\h; \bbtheta)$ is our optimization variable, which we compare with $\Dx^*(\h)$ through their KL divergence. KL divergences of different channel realizations are averaged over the channel state distribution $\Dh$, which is also given. The optimal distribution $\Dx^*(\h; \bbtheta) = \Dx^*(\x \given \h; \bbtheta)$ minimizes the expected KL divergence among those representable by the parametric family $\Dx(\h; \bbtheta)$.

To solve \eqref{problem:supervised-diffusion-learning}, we need access to the expert conditional distributions $\Dx^*(\h)$. This is impossible in general because algorithms that solve \eqref{problem:stochastic-functional} do not solve for $\Dx^*(\h)$ directly. Rather, algorithms that solve \eqref{problem:stochastic-functional} generate samples $\x(\h)$ drawn from the optimal distribution $\Dx^*(\h)$ \cite{StateAugmented_RRM_GNN_naderializadeh_TSP2022}. Thus, we recast the goal of this paper as learning to generate samples  $\x \given \h$ from the distribution $\Dx^*(\h; \bbtheta)$ when we are given samples $\x(\h)$ of the expert conditional distributions $\Dx^*(\h)$ with channel states generated according to $\Dh$:

\medskip

\begin{itemize}

\item[] \begin{problem}\label{the_problem} Given samples $\x(\h)$ drawn from the expert  distribution $\Dx^*(\h) \Dh = \Dx^*(\bbx\given\h)\Dh$ [cf. \eqref{problem:stochastic-functional}], we learn to generate samples $\x \given \h$ drawn from the conditional distributions $\Dx^*(\h; \bbtheta) = \Dx^* (\x \given \h; \bbtheta)$ [cf. \eqref{problem:supervised-diffusion-learning}]. \end{problem}

\end{itemize}

\medskip

\noindent A solution of Problem \ref{the_problem} is illustrated in Fig.~\ref{fig:example-generations}. For a given channel state realization $\h$, we show two-dimensional slices of \emph{samples} of an optimal policy (in blue). As indicated by \eqref{fig:global-opt-metrics}, these samples realize optimal QoS metrics for \eqref{problem:stochastic-functional} if executed sequentially (Fig.~\ref{fig:global-opt-metrics}). We train a generative diffusion model (Section \ref{sec:proposed-gdm-policies}) that generates samples (in orange) that are distributed close to samples of an optimal distribution. When executed sequentially, the learned samples realize QoS metrics close to optimal values (Fig.~\ref{fig:global-opt-metrics}). We underscore that neither the optimal distribution $\Dx^*(\h) \Dh = \Dx^*(\bbx\given\h)\Dh$ nor the parametric distribution $\Dx^*(\h; \bbtheta) = \Dx^* (\x \given \h; \bbtheta)$ is computed.

\subsection{Learning in the Dual Domain \& Policy Randomization}
Most learning approaches to allocating resources in wireless systems contend with the deterministic policy formulation in \eqref{problem:deterministic-functional}, e.g., {\cite{wang2022decentralized, liang2019towards, sun2017learning, Du2024EnhancingOptimization, diffsg2024liang, Liang2024DiffusionModelsNetworkOptimizers, Lu25Energy, darabi2024diffusion}}. This is due in part to the use of deterministic learning parameterizations {\cite{wang2022decentralized, liang2019towards, sun2017learning}} but even recent contributions that propose diffusion models do so for deterministic policies {\cite{Du2024EnhancingOptimization, diffsg2024liang, Liang2024DiffusionModelsNetworkOptimizers, Lu25Energy, darabi2024diffusion, xue2025joint}}.
This is a well-known limitation that has motivated, e.g., state-augmented algorithms that leverage dual gradient descent dynamics to randomize policy samples \cite{StateAugmented_RRM_GNN_naderializadeh_TSP2022, uslu2025faststateaugmentedlearningwireless}. These algorithms generate trajectories of primal and dual iterates by operating on a convex hull relaxation of the Lagrangian for the original problem and iteratively solving a sequence of Lagrangian maximization subproblems. Each subproblem is an unconstrained, deterministic problem to which regular learning methods apply, and near-optimality and feasibility guarantees are established neither for individual primal iterates nor their averages, but only for the sequential execution of the generated policy iterates.

A shortcoming of state-augmented algorithms---and iterative dual domain algorithms in general---is that they incur a transient period where suboptimal policies are executed. Reducing the length of this transient period typically requires larger step sizes, which in turn introduces a trade-off with respect to solution optimality. Learning a generative model to sample from the stationary (optimal) policy distribution emerges as a promising approach for overcoming this trade-off. To the best of our knowledge, our paper is the first to develop and demonstrate imitation of stochastic policies that solve a constrained optimization problem with generative diffusion models.

\section{Policy Generative Models}
\label{sec:proposed-gdm-policies}

GDMs involve a forward and a backward diffusion process. The \emph{forward} process defines a Markov chain of diffusion steps to progressively add random noise to data. For a given $\h$ and a data sample $\x_0 = \x(\h)$ drawn from the expert distribution $\Dx^\star(\h)$, the forward chain follows
\begin{align} \label{eq:forward-transition}
    q(\bbx_k \cond \bbx_0; \h) = \mathcal{N}(\bbx_k; \sqrt{\bar{\alpha}_k} \bbx_0, (1 - \bar{\alpha}_k) \bbI),
\end{align}
where $\bar{\alpha}_k := \prod_{i=1}^k \alpha_i$, $\alpha_k := 1 - \beta_k$, and $\beta_k$ is a monotonically increasing noise schedule, e.g., linear. For a sufficiently large $K$,~\eqref{eq:forward-transition} converts the data sample $\x_0$ into a sample that is approximately isotropic Gaussian distributed, i.e., $\bbx_K \approx \mathcal{N}(\bb0, \bbI)$.

The reverse process of~\eqref{eq:forward-transition} is approximated by a chain of Gaussian transitions with a parametrized mean $\bbmu_{\bbtheta}$ and fixed variance $\sigma^2_k \bbI$,
\begin{align} \label{eq:reverse-mean-estimator}
    p_{\bbtheta}(\bbx_{k-1} \cond \bbx_k; \h) = \mathcal{N} \big(\bbx_{k-1}; \bbmu_{\bbtheta}(\bbx_k, k; \h), \sigma^2_k \bbI \big).
\end{align}

A \emph{backward} diffusion process samples $\bbx_K \sim \mathcal{N}(\bb0, \bbI)$ and iteratively runs the backward chain in \eqref{eq:reverse-mean-estimator} for $k = K, \ldots, 1$. With reparametrization of~\eqref{eq:forward-transition} as $\bbx_k(\bbx_0, \bbepsilon) = \sqrt{\bar{\alpha}_k} \bbx_0 + \sqrt{1 - \bar{\alpha}_k} \bbepsilon$ \cite{ho2020denoising}, sampling $\bbx_{k-1} \sim p_{\bbtheta}(. \cond \bbx_k; \h)$ amounts to updating
\begin{align}
\label{eq:diffusion-sampling}
    \bbx_{k-1} = \frac{1}{\sqrt{\alpha}_k} \left( \bbx_k - \frac{\beta_k}{\sqrt{1 - \bar{\alpha}_k}} \bbepsilon_{\bbtheta}(\bbx_k, k; \h)  \right) + \sigma_k \bbw, 
\end{align}
where $\bbw \sim \mathcal{N}(\bb0, \bbI)$, and $\bbepsilon_{\bbtheta}(
\bbx_k, k; \h)$ predicts the noise $\bbepsilon$ added to $\bbx_0 \sim \Dx^\star(\h)$ from noisy sample $\bbx_k$ at timestep $k$.

An \emph{optimal GDM-policy parametrization} $\bbtheta^\star$ minimizes the $\h$-expectation of the DDPM loss function \cite{ho2020denoising} given by
\begin{align} \label{eq:gdm-loss-statistical}
    \Lgdm(\bbtheta) = \E_{\bbx_0, k, \h, \bbepsilon} \, \omega(k) \!\left \| \bbepsilon_{\bbtheta} \big( \bbx_k(\bbx_0, \bbepsilon), k; \h \big) - \bbepsilon \right \|^2\!.
\end{align} 
\noindent In~\eqref{eq:gdm-loss-statistical}, $\omega(k)$ is a time-dependent weighting function
and the expectation is over random timesteps $k \sim \mathrm{Uniform}([1, K])$, Gaussian noise $\bbepsilon \sim \mathcal{N}\left( \bb0, \bbI \right)$, expert (data) samples $\bbx_0 \sim \Dx^\star(\h)$, and conditioning networks $\h \sim \Dh$. 

We note that the DDPM loss in~\eqref{eq:gdm-loss-statistical} is a variational upper bound on the expected KL divergence loss in \eqref{problem:supervised-diffusion-learning}, which becomes tight when $\bbtheta = \bbtheta^\star$. Thus, running~\eqref{eq:diffusion-sampling} with optimal parametrization $\bbepsilon_{\bbtheta^\star}$ for a given $\h$ generates samples from the expert conditional distribution, i.e., $\bbx_0 \sim \Dx(\h; \bbtheta^\star) = \Dx^\star(\h; \bbtheta) \approx \Dx^\star(\h)$.

\vspace{1em}
\section{GNN-Parametrizations for GDM Policies}
\label{sec:gnn-architectures}

We employ GNNs for GDM parameterization, as they are well-suited for processing network data, such as resource allocations. Moreover, GNNs inherently take graphs as input, making them a natural fit for GDMs conditioned on $\h$. 

GNNs process graph data through a cascade of $L$ graph convolutional network (GCN) layers \cite{gama2018convolutional}. Inputs are node signals (features) and graph shift operators (GSO) while outputs are node embeddings. Each GCN layer $\bbPsi^{(\ell)}$ is a nonlinear aggregation function obtained by the composition of a graph convolutional filter and a pointwise nonlinearity $\varphi$ (e.g., relu), 
\begin{align} \label{eq:graph-conv-layer}
\hspace{-.7em}\bbZ^{(\ell)} \!=\! \bbPsi^{(\ell)}\left(\bbZ^{(\ell-1)}; \bbH, \Theta^{(\ell)}\right) \!=\! \varphi  \left[ \,  \sum_{m = 0}^{M_{\ell}} \;  \bbH^m \bbZ^{(\ell-1)} \bbTheta_{m}^{(\ell)}  \right]\!.
\end{align}
\noindent In \eqref{eq:graph-conv-layer}, $\Theta^{(\ell)} = \{ \bbTheta_k^{(\ell)} \in \reals^{F_{\ell-1} \times F_\ell} \}_{k = 0}^{K}$ is a set of learnable weights, $M_{\ell}$ denotes the number of hops, and $\bbZ^{(\ell-1)} \in \reals^{N \times F_{\ell-1}}$ is the input node signal to layer $\ell$. The GSO, $\h$, encodes the underlying connectivity of the network, which is the network state in our case. 

For improved and more stable training, we take advantage of normalization layers and residual connections. To this end, we redefine $\varphi$ in \eqref{eq:graph-conv-layer} as the composition of a normalization layer followed by a pointwise nonlinearity, while the first term in the sum, $\bbZ \bbTheta_0$, inherently represents a learnable residual connection.

We view $\bbx_k$ and $\bbk = k \mathbf{1}_{N}$ as node signals and 
introduce a read-in layer $\bbPhi^{(0)} = (\bbPhi_{\bbx}, \bbPhi_{\bbt})$ that adds sinusoidal-time embeddings to the input node features. That is, we have
\begin{align} \label{eq:read-in-layer}
    \bbZ^{(0)} = \bbPhi^{(0)}(\bbx_k, \bbk) = \bbPhi_{\bbx} (\bbx_k) + \bbPhi_{\bbk} (\bbk),
\end{align}
where $\bbPhi_{\bbx}: \reals^{N} \mapsto \reals^{N \times F_0}$ is a multilayer perceptron (MLP) layer, and  $\bbPhi_{\bbt}: \reals^N \mapsto \reals^{N \times F_0}$ is a cascade of a sinusoidal time embedding and MLP layers. Finally, we add a readout MLP layer $\bbPhi^{(L)}: \reals^{N \times F_{L}} \mapsto \reals^N$ that learns to predict the noise $\bbepsilon$ from the output node embeddings. 

\section{Case Study: Power Control in Multi-User Interference Networks}
\label{sec:experiments}

\begin{figure*}[ht!]
    \centering
    \begin{minipage}{.33\linewidth}
    \includegraphics[width=\linewidth]
    {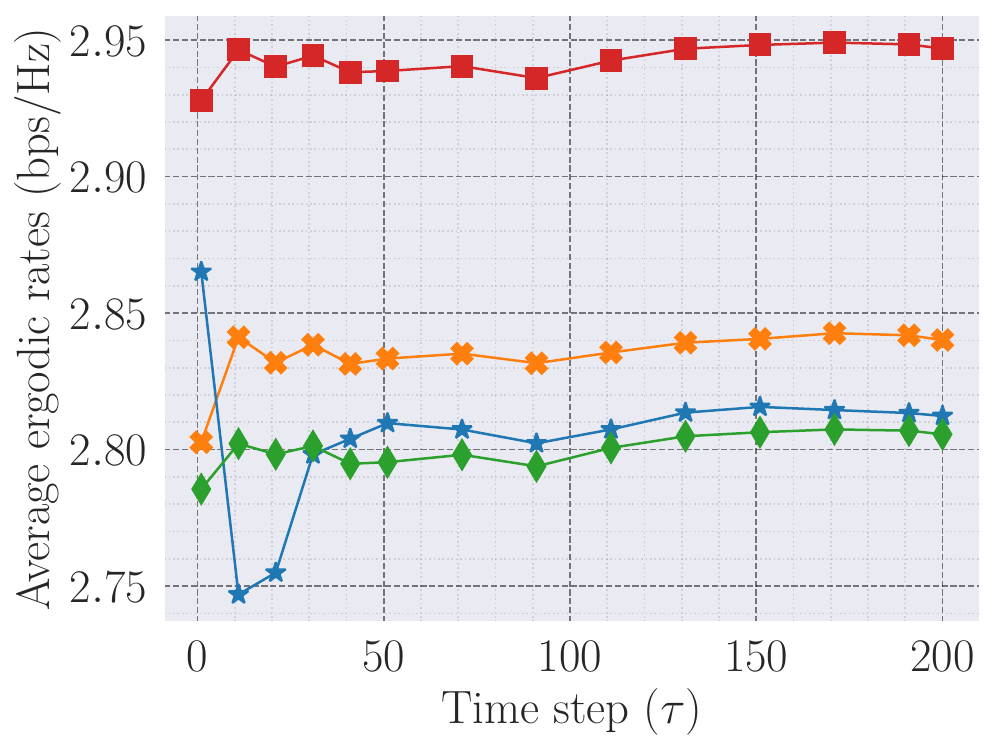}
    \end{minipage}%
    \hfill
    \begin{minipage}{.33\linewidth}
    \vspace{-2.7em}
        \includegraphics[width=\linewidth]
        {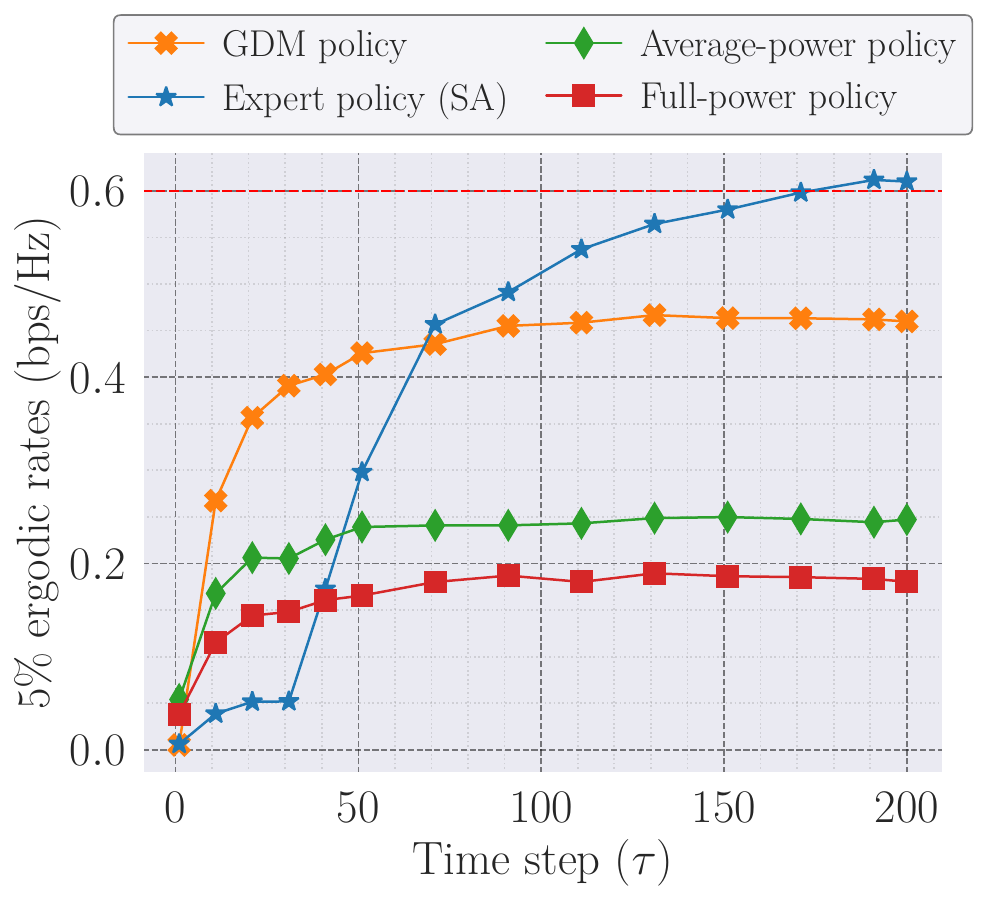}
        
    \end{minipage}
    \hfill 
    \centering
    \begin{minipage}{.33\linewidth}
    \vspace{-2em}
        \includegraphics[width=\linewidth]
        {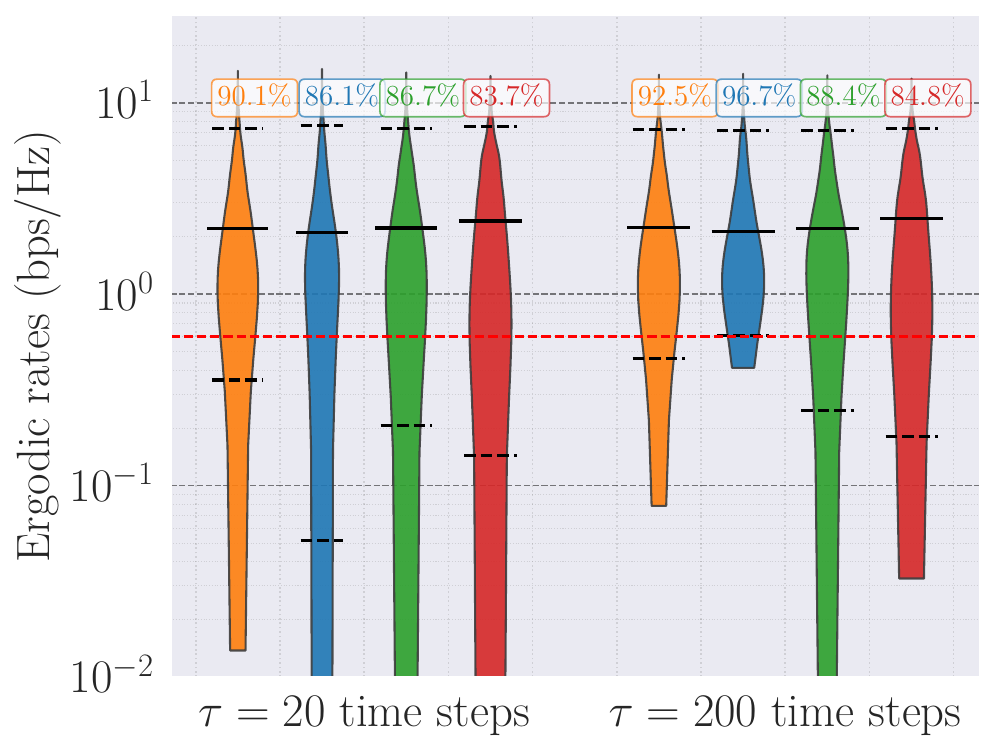}
    \end{minipage}%
    \hfill 
    \caption{Comparison of the test performance of GDM policy with the expert policy (SA) and other baselines. Leftmost and middle plots show the time evolution of the average and 5th percentile of ergodic rates, respectively. The minimum rate requirement $f_{\min}$ is shown with a dashed, red line. Rightmost plot shows the distribution of ergodic rates and constraint satisfaction percentages, evaluated up to $\tau = 20$ and $\tau = 200$ time steps. Solid black lines are drawn at the median values while the dashed black lines are for the 5th and 95th percentile values.}
    \label{fig:global-opt-metrics}
\end{figure*}

We consider the problem of power control in $N$-user interference channels. Similar setups have been investigated in~\cite{StateAugmented_RRM_GNN_naderializadeh_TSP2022, uslu2025faststateaugmentedlearningwireless} and should be referred to for more details. 

\subsection{Wireless Network \& Power Control Setup}

To summarize the setup briefly, all network realizations are sampled from a family of network configurations with $N = 100$ transmitters-receiver (tx-rx) pairs (also nodes in our graphs) and an average density of $12$ tx-rx pairs/\SI{}{\kilo\meter\squared}. For each network, we first drop the transmitters randomly in a square grid world, and each transmitter is paired with a neighboring receiver. Signals coming from all but their respective transmitters are treated as interference by the receivers.

We optimize the transmit power levels $\x \in [0, P_{\max}]^{N}$ 
where $P_{\max} = 10$ mW is the maximum transmit power budget. The channel bandwidth and noise power spectral density (PSD) are set to $W = 20$ MHz and $N_0=-174$ dBm/Hz, respectively. The network state $\h$ is the matrix of long-term channel gains which follow a log-normal shadowing with a standard deviation of $7$ plus the standard dual-slope path-loss model. For a given $\h$, the short-term (instantaneous) channel gains $\widetilde{\h}$ vary following Rayleigh fading. To evaluate the performance of a policy $\bbx$, we define the instantaneous rate of receiver $i$ as
\begin{align} \label{eq:rate-instantaneous}
\widetilde{r}_i(\x, \widetilde{\h}) = \log_2 \left( 1 + \frac{x_i \cdot |\widetilde{h}_{ii}|^2}{W N_0 + \sum_{j\neq i} x_j \cdot |\widetilde{h}_{ji}|^2}  \right),
\end{align}
where $x_i$ is the $i$th component of $\bbx$ and $\widetilde{h}_{ji}$ is the $(j,i)$th element in matrix $\widetilde\h$. Observe that a policy $\Dx(\h)$ is determined only by the long-term gains, whereas the instantaneous rate depends on the short-term channel gains. Ergodic rates are defined as
\begin{align} \label{eq:rate-ergodic}
\bbr(\Dx(\h), \h) := \E_{\Dx(\h), \; \widetilde{\h} \vert \h} \left[ \widetilde{\bbr} \big( \x, \widetilde{\h} \big) \right].
\end{align}
\noindent In \eqref{eq:rate-ergodic}, we take an expectation over the policy and the fading jointly. In our experiments, we draw $200$ samples from the trained GDM policy for each network and evaluate the joint expectation over 200 time steps, with each time step spanning 10 ms.

A minimum ergodic rate requirement of $f_{\min} = 0.6$ bps/Hz is imposed for all receivers by setting the utility constraints $\bbf(\Dx(\h), \h) := \bbr (\Dx(\h), \h) - \mathbf{1}_{N} f_{\min}$ whereas the utility objective is the network-wide average of the ergodic rates given by $f_0(\Dx(\h), \h) := \mathbf{1}^\top_{N} \bbr(\Dx(\h), \h) / N$.

\subsection{State-Augmented Primal-Dual Expert Policy \& Baselines}
To generate samples from an optimal solution distribution, we first train a GNN-parametrized model via a state-augmented primal-dual (SA) learning algorithm as in \cite{uslu2025faststateaugmentedlearningwireless}. The trained model is executed online for each given network $\h$ over a sufficiently long time window to generate trajectories of primal and dual iterates with near-optimality and feasibility guarantees. We collect the resulting primal iterates $\{ \bbx^\dagger_{b}(\h) \}_{b = 0}^{B-1}$, i.e., resource allocation vectors, in a buffer with capacity $B = 500$. During GDM policy training, expert policy samples are uniformly drawn from the buffer, i.e., we have $\Dx^\star(\h) = \mathrm{Uniform} \big[ \{ \bbx^\dagger_0(\h), \ldots, \bbx^\dagger_{B-1}(\h) \} \big]$ for all $\h \sim \Dh$. Note that while we parametrize the expert policy and run a state-augmented training algorithm over a training dataset of networks $\Dh$, one can alternatively run the dual descent without the parametrization and store the resource allocation iterates for each instance $\h \sim \Dh$. 

We compare the GDM and expert policies with two baseline deterministic policies:
\begin{enumerate}
    \item \emph{Average-power transmission policy}: For each given $\h$, we compute the time-average of the expert policy samples--similar to primal averaging in the optimization literature--and fix it, i.e., we set $\bbx(\h) \approx \E_{\Dx^\star(\h)} [\bbx^\dagger(\h)]$ at all time steps.

    \item \emph{Full-power transmission policy (FP)}:  All transmitters use all the transmission power available at all time steps, i.e., $\bbx(\h) = P_{\max} \mathbf{1}_N$.
\end{enumerate}

\subsection{Implementation \& Training Details for GDM Policy}
For the GNN-parametrizations, the GSO representation of the network state $\h$ is a fully connected graph where nodes correspond to transmitter-receiver pairs, and the edge weights between nodes $i$ and $j$, $e_{ij}$, are set to log-normalized long-term channel gains given by $\displaystyle e_{ij} \propto \log_2 \left( 1 + \frac{P_{\max} |h_{ij}|^2}{WN_0} \right)$. The GNN has $L = 6$ layers, each with $F_{\ell}= 128$ hidden features and $M_{\ell} = 2$ hops (filter taps). We set the number of diffusion time steps to $K=500$, use a cosine noise schedule $\beta_k$ \cite{nichol2021cosine} and train the GDM policy to minimize the DDPM objective in \eqref{eq:gdm-loss-statistical} with a log-SNR weighting function $\omega(k) = \log \big( \mathrm{SNR}(k) \big)$, where $\mathrm{SNR}(k) := \alpha^2_k / \sigma^2_k$. 

To evaluate our method, we draw a total of $128$ network realizations. Following a $5:1:2$ split, we obtain training, validation, and test datasets of size $|\Dh| = 80$, $|\ccalV_{\h}| = 16$, and $|\ccalT_{\h}| = 32$ networks, respectively. The expert policy solutions are obtained across all three datasets for evaluation and benchmarking purposes. We train the GDM policy over the training dataset $\Dh$ for a maximum of $10^4$ epochs with an ADAM optimizer \cite{kingma2014adam}, an initial learning rate of $10^{-2}$, and a learning rate schedule that follows a cosine decay with warm restarts. In each epoch, we iterate over the whole dataset with mini-batches of $16$ graphs and sample $250$ graph signals, i.e., expert policy data $\bbx(\h) \sim \Dx^\star(\h)$, for each graph in the batch. Every $200$ epochs, we evaluate the checkpointed GDM model on the validation dataset $\ccalV_{\h}$ and save the best model in terms of $5$th percentile ergodic rates. We test the saved model on $\ccalT_{\h}$.     

We apply an affine transform $[0, P_{\max}] \mapsto [-1/2, 1/2]$ to map the policy space to a centered diffusion space. To sample from the diffusion model, we run the DDPM sampling equation~\eqref{eq:diffusion-sampling} with standardized variables, invert the affine transform, and project the generated policy samples to the support $[0, P_{\max}]^N$.\footnote{During inference, we observed negligible difference in the quality of generations when we swapped the DDPM sampler with other samplers, e.g., a DDIM sampler or its accelerated counterpart with fewer denoising time steps.}

\begin{figure}[t!]
    \centering
    \begin{minipage}{.49\linewidth}
    \includegraphics[width=\linewidth]{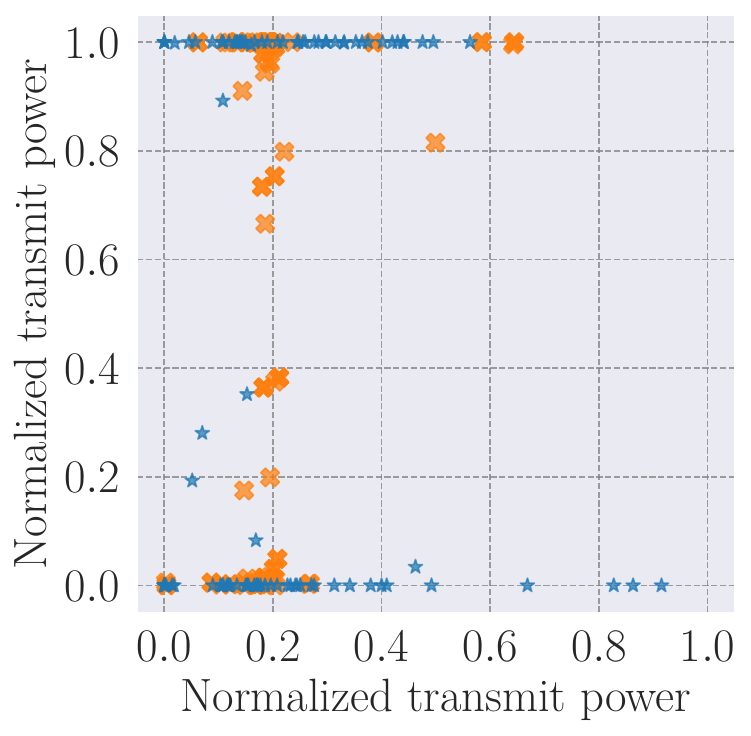}
    \end{minipage}%
    \hfill
    \begin{minipage}{.49\linewidth}
        \includegraphics[width=\linewidth]
        {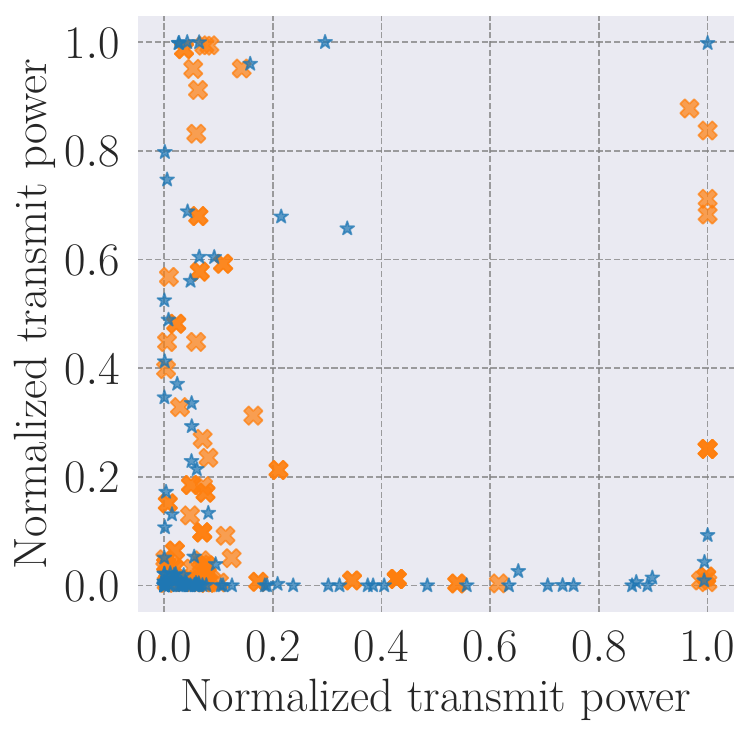}
    \end{minipage}
    \hfill 
    \vfill 
    \centering
    \begin{minipage}{.49\linewidth}
    \includegraphics[width=\linewidth]{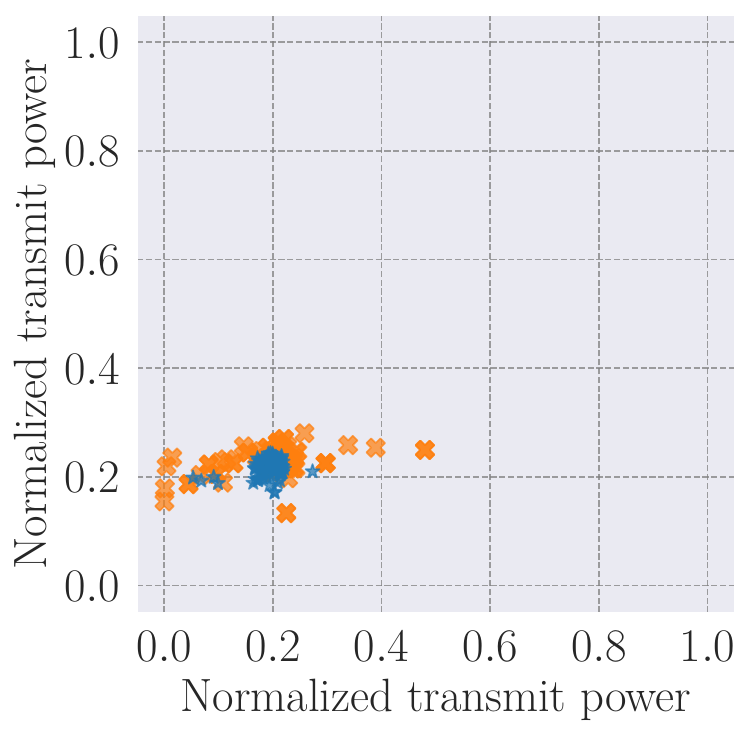}
    \end{minipage}%
    \hfill
    \begin{minipage}{.49\linewidth}
        \includegraphics[width=\linewidth]
        {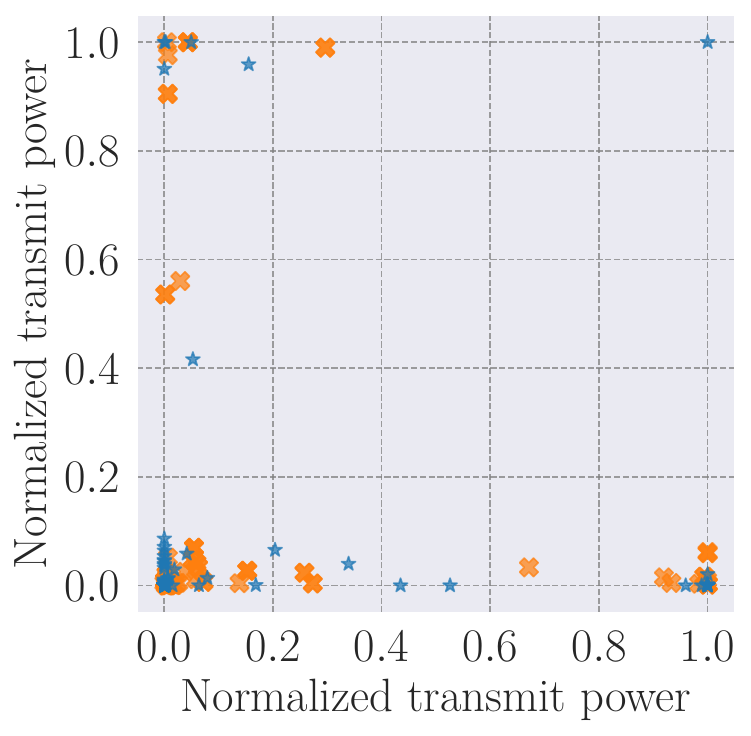}
    \end{minipage}
    \hfill 
    \caption{Example two dimensional slices of expert policy (blue) and GDM policy (orange) samples are shown for two pairs of neighboring nodes from the test dataset. Transmit powers are normalized by $P_{\max}$.
    }
    \label{fig:example-generations}
\end{figure}


\subsection{Performance of GNN-Parametrized GDM Policy}
Fig.~\ref{fig:global-opt-metrics} showcases the test performance of the GDM policy, expert policy, and the baselines over a time horizon of $200$ time steps. We estimate the ergodic rate vector for a given network $\h$ and time step $\tau$ as $(1/\tau) \sum_{s = 0}^{\tau - 1} \widetilde{\bbr} \big( \bbx_{s}, \widetilde{\h} \big)$ where $\bbx_s$ denotes the power allocation decision at an earlier time step $s < \tau$ [c.f.~\eqref{eq:rate-ergodic}].  In the first two plots, we report the time evolution of the average ergodic rate, i.e., the mean-rate objective utility, and the 5th-percentile of ergodic rates, both of which are computed across all $| \ccalT_{h} | \times N = 32 \times 100 = 3200$ tx-rx pairs in the test dataset. The rightmost violin plot shows the histogram of ergodic rates of all receivers evaluated at two different time steps. We verify that the expert policy eventually satisfies almost all the constraints unlike the baselines. Although the GDM policy does not exhibit strict feasibility, it converges to a near-feasible and near-optimal policy in very few time steps compared to the expert policy and does not incur a long transient period. Moreover, both the full-power and average-power baselines are outperformed by the GDM policy in terms of percentile rates and constraint violations.

In Fig.~\ref{fig:example-generations}, example two-dimensional (2D) slices of $100$-dimensional learned GDM policies are overlaid with expert policy samples. The GDM policy generalizes to unseen test networks drawn from $\ccalT_{\h}$, and the conditional distribution of GDM policy samples significantly resembles that of the expert policy. A peculiarity of optimal power control policies is that they tend to be probabilistic and involve multiple transmission modes. That is especially true for tx-rx pairs with less favorable channel conditions for which the policy randomization becomes more nuanced. In such cases, similar to time-sharing strategies, pairs that would otherwise generate considerable mutual interference adopt a policy-switching mechanism where they take turns to transmit at high power during periods of minimal interference (e.g., the top-left and bottom-right corners in the rightmost plot of Fig.~\ref{fig:example-generations}). By alternating their transmissions, all tx-rx pairs satisfy their minimum ergodic rate requirements. 

Strong test generalization evidenced in both figures notwithstanding, the GDM policy exhibits a small feasibility gap compared to the expert policy in Fig.~\ref{fig:global-opt-metrics}. We attribute this gap primarily to the supervised training algorithm not accounting directly for the sensitivity of the constraints and the aforementioned policy-switching phenomenon. The feasibility gap and overall performance of the GDM policies can be further improved by incorporating the QoS requirements and additional variance constraints directly into the training loss and/or generative process.

\section{Conclusion}\label{sec:conclusion}
This work demonstrated that generative diffusion processes can imitate expert policies that sample from optimal solution distributions of stochastic network optimization problems. We employed a GNN to condition the generative process on a family of wireless network graphs. More broadly, we anticipate that our attempt at generative diffusion-based sampling of random graph signals will be of interest beyond resource optimization in wireless networks. We leave constraint-aware, unsupervised training of GDM policies across a wider range of network topologies as future research directions.


\newpage
\clearpage
\bibliographystyle{IEEEbib}
\bibliography{./bibliography/refs
}

\end{document}